\documentclass[letterpaper, 10 pt, journal, twoside]{ieeetran}
\usepackage{amsmath,amsfonts}
\usepackage{algorithmic}
\usepackage{algorithm}
\usepackage{array}
\usepackage[caption=false,font=normalsize,labelfont=sf,textfont=sf]{subfig}
\usepackage{textcomp}
\usepackage{stfloats}
\usepackage{url}
\usepackage{verbatim}
\usepackage{graphicx}
\usepackage{cite}
\usepackage{siunitx}
\usepackage{multirow}
\usepackage{booktabs}
\usepackage{hyperref}

\begin{document}

\title{ORION: Option-Regularized Deep Reinforcement Learning for Cooperative Multi-Agent Online Navigation}

\author{Shizhe Zhang$^{*}$, Jingsong Liang$^{*}$, Zhitao Zhou, Shuhan Ye, Yizhuo Wang, Derek Ming Siang Tan, \\Jimmy Chiun, Yuhong Cao, and Guillaume Sartoretti,~\IEEEmembership{Member,~IEEE}

\thanks{Manuscript received December 30, 2025; Revised February 24, 2026; Accepted May 4, 2026. This paper was recommended for publication by Editor M. Ani Hsieh upon evaluation of the Associate Editor and Reviewers’ comments. This work was supported by Temasek Laboratories (TL@NUS) under grant TL/FS/2025/01. \textit{(Corresponding author: Yuhong Cao.)}}
\thanks{$^{*}$ Equal Contribution. The authors are with the Department of Mechanical Engineering, College of Design and Engineering, National University of Singapore, Singapore 117575 (email: \tt\footnotesize $\{$shizhezhang,jingsongliang,zhitao\_zhou,e1373216,wy98,\\derektan,jimmy.chiun,caoyuhong$\}$@u.nus.edu; mpegas@nus.edu.sg).}%
\thanks{The code is available at \url{https://github.com/marmotlab/ORION}.}
}

\markboth{IEEE Robotics and Automation Letters. Preprint Version. Accepted May, 2026}
{Zhang \MakeLowercase{\textit{et al.}}: Option-Regularized Deep Reinforcement Learning for Cooperative Multi-Agent Online Navigation} 

\maketitle

\begin{abstract}
Existing methods for multi-agent navigation typically assume fully known environments, offering limited support for partially known scenarios with outdated or imperfect prior maps, such as warehouses or factory floors. There, agents need to balance path optimality with collecting and sharing environmental information to help teammates reach their own targets. To these ends, we propose \textit{ORION}, a novel deep reinforcement learning framework for cooperative multi-agent online navigation in partially known environments. Starting from an imperfect prior map, ORION trains agents to make decentralized decisions, coordinate toward individual targets, and actively reduce task-relevant map uncertainty through online observation sharing in a closed perception–-action loop. We first design a shared graph encoder that fuses prior map with online perception into a unified representation, providing robust state embeddings under environmental discrepancies. At the core of ORION is an option-critic framework that learns high-level cooperative modes translated into sequences of low-level actions, enabling adaptive switching between individual navigation and team-level exploration. We further introduce a dual-stage cooperation strategy that allows agents to assist teammates under map uncertainty, thereby reducing the overall makespan. Across extensive maze-like maps and large-scale warehouse environments, ORION achieves high-quality real-time decentralized cooperation while scaling to up to 10 robots, outperforming state-of-the-art classical and learning-based baselines. Finally, we validate ORION on physical robot teams, demonstrating its robustness and practicality for real-world cooperative navigation.
\end{abstract}


\section{Introduction}
\IEEEPARstart{M}{ulti-agent} navigation is a fundamental robotic problem, where a team of agents must reach individual targets while minimizing the overall makespan, i.e., the travel distance or time until the last agent reaches its assigned target. Beyond the challenges of single-agent navigation, such as long-horizon planning and dynamic obstacle avoidance, multi-agent navigation further introduces team-level cooperation: agents need to anticipate/react to dynamic interactions arising from neighboring robots and changes in shared environments. However, many real-world deployments, especially in warehouses or factory floors, operate under imperfect or outdated prior maps due to frequent layout changes and occlusions. This motivates a nonstandard yet practical variant of multi-agent navigation, where robots must plan and coordinate online while jointly reducing map uncertainty through decentralized exploration and information sharing. This setting amplifies two key challenges: balancing individual target-reaching with team-level information gathering, and reasoning about dynamic multi-robot interactions under partial observability.
In such cases, robots must navigate while continuously updating their belief of the map, requiring richer team-level reasoning that goes beyond classical (sometimes heuristic and offline) path planning. That is, environmental uncertainty forces the team to adaptively trade off between exploiting known areas and exploring unknown areas that may benefit the team's overall objective (e.g., find shortcuts for other agents, or help them avoid dead-ends). Cooperation is therefore more complex: a robot may need to temporarily deviate from its shortest path or intentionally gather information that primarily benefits others, sometimes at the expense of its own budget.

\begin{figure}[t]
  \centering
  \includegraphics[width=0.9\linewidth]{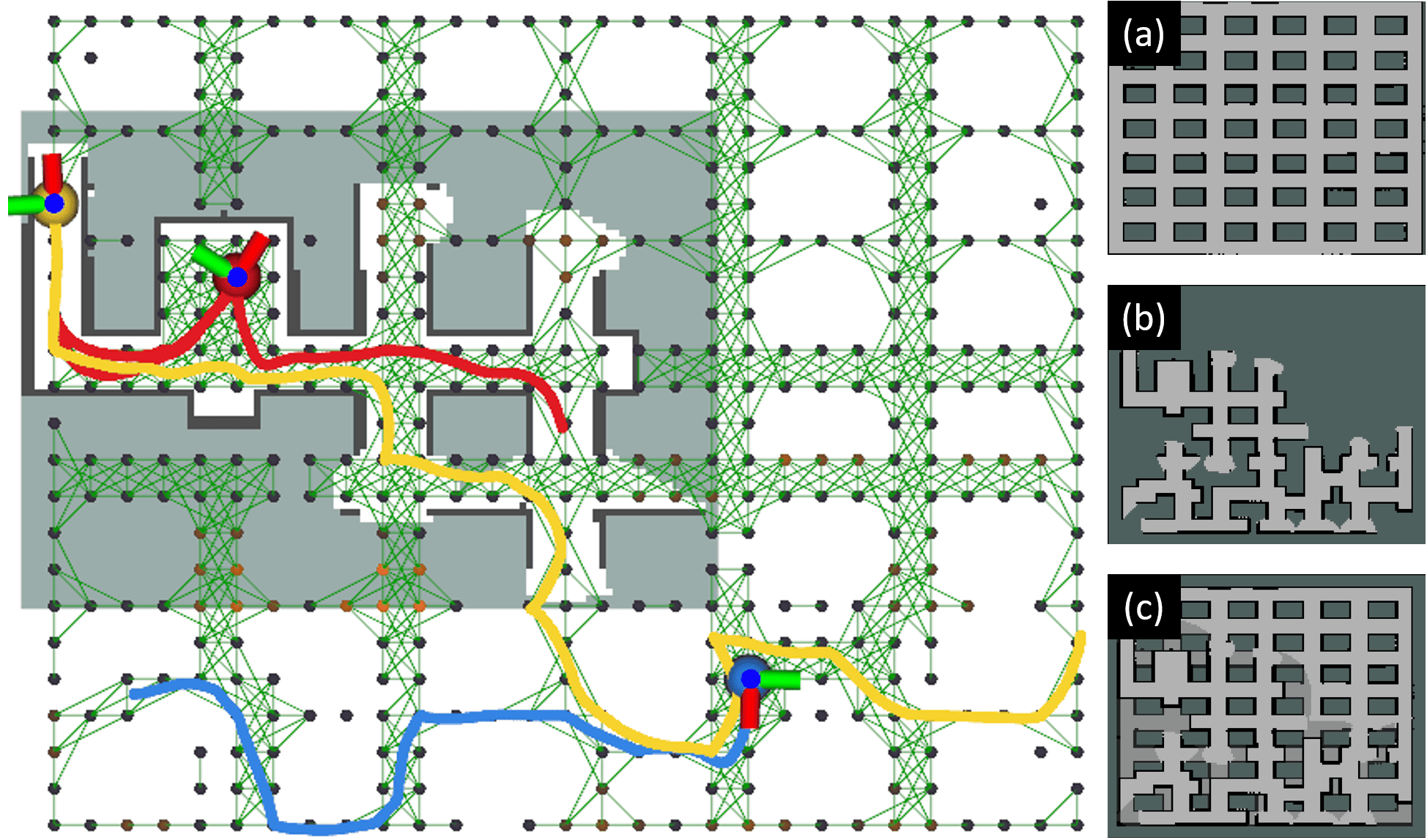}
  \caption{Overview of Multi-Agent Online Navigation. Each agent is assigned a target and navigates over a prior map that may differ from the ground truth. Agents maintain/share (a) \emph{prior map}, (b) \emph{current map}, and (c) \emph{combined map} that fuse prior/online sources to reason about partially changed environments. During navigation, agents not only pursue their own targets but also cooperate by sharing information and assisting others. For example, the red agent reaches its target early and then helps the yellow agent by exploring uncertain regions before returning. ORION enables such adaptive cooperation both before and after arrival on-goal, ultimately reducing the team's makespan by coordinating agents to contribute where they are most needed in a decentralized way.}
  \label{fig: attention vis}
  \vspace{-0.2cm}
\end{figure}

Early works in multi-agent navigation primarily focused on ensuring collision-free motion at the kinematic level, typically by decoupling target-driven planning from local obstacle/neighboring robots avoidance~\cite{jin2021hierarchical,yu2023learning,wang2020mobile}. These approaches largely treat interactions among the robot team as safety constraints rather than opportunities to enhance team-level performance: robots rarely leverage team-level coordination and shared global information to guide long-horizon decision-making. Another line of work studies formation-preserving or centroid-driven navigation, where coordinated motions yield more globally consistent behaviors, e.g., maintaining formation structure and avoiding oscillations/conflicts~\cite{tanner2012multiagent,quan2023robust}. Yet, these approaches typically assume a known environment or fixed/specific formation objectives, which limits their flexibility in open-ended navigation tasks where robots must dynamically adapt their behaviors to evolving map information. Another closely related domain is Multi-Agent Path Finding (MAPF), where the goal is to compute globally coordinated, conflict-free trajectories for all robots (usually offline), often emphasizing scalability in large structured environments~\cite{li2021eecbs,li2022mapf,sartoretti2019primal}. Despite the global coordination achieved by MAPF approaches, they usually assume a fully known environment and lack the ability to reason about partial observability, exploration, or online map updates.%

To address these challenges, we propose \textit{ORION}, a novel deep reinforcement learning-based framework designed for cooperative multi-agent online navigation. ORION first introduces a fused graph-based representation of each agent's prior and online updated maps, enabling the downstream policy network to jointly reason about known structures and newly observed areas. We further propose an option-critic structure in which an \textit{option} denotes a high-level policy that either focuses on the agent's individual navigation or on supporting teammates by collecting team-level informative observations about the environment. Each option persists for multiple time steps and provides the corresponding low-level action guidance. A learned termination mechanism then decides, based on local observations and the team's state, whether to continue the current option or switch to a more suitable one. Embedding this option mechanism into both the policy and critic networks enables ORION to judge/evaluate the long-term impact of option sequencing, allowing robots to adopt behaviors that balance individual progress with team-level gains. Within this option framework, we further design a two-stage cooperation strategy that captures distinct behaviors before and after an agent reaches its target. Before arrival, agents navigate toward their targets while opportunistically gathering information that can reduce future detours and alleviate team-level information gaps. After arrival, instead of remaining idle, agents may choose to explore nearby areas to assist other agents still en route, adaptively deciding how much additional exploration is beneficial without jeopardizing their timeliness. This strategy is validated in our experiments to benefit the makespan by resolving remaining uncertainties for agents still ``suffering'' from individual exploration-exploitation trade-offs. We evaluate ORION extensively across hundreds of simulated maps, large-scale Gazebo environments, and real-world deployments with varying team sizes and start–target assignments. ORION consistently outperforms state-of-the-art classical and learning-based planners by \SI{6.9}{\%}-\SI{13.4}{\%} in makespan while maintaining real-time inference efficiency. We further demonstrate up to an additional \SI{14.2}{\%} improvement arising specifically from our two-stage cooperation strategy. Finally, real-world experiments further demonstrate that ORION transfers without additional training/tuning, highlighting its practicality for cooperative navigation under partial observability.

\section{Related Work}
\subsection{Navigation in Partially Known Environments}
Navigation in fully known environments has been extensively studied, with search-based planners relying on efficient heuristic-guided graph expansion and sampling-based planners exploring the continuous space through incremental sampling~\cite{koenig2005fast,likhachev2005anytime,karaman2011sampling,gammell2015batch}. While these methods are effective in static and fully known settings, they tend to degrade in partially known environments due to their limited ability to anticipate unobserved environment and the high computational cost of frequent replanning.

Learning-based navigation methods, particularly deep reinforcement learning, have shown promise in reasoning over uncertainty without hand-crafted heuristics. Many of these approaches, however, are designed primarily for local/motion-level planning with sensory inputs such as RGB-D images or point clouds~\cite{pfeiffer2017perception,jin2020mapless}, making them difficult to scale to large, complex environments where long-horizon reasoning is essential~\cite{tai2017virtual}. Several approaches employ belief-informed~\cite {liang2023context} or hierarchical policies~\cite{liang2024hdplanner} to enhance spatial reasoning and long-term waypoint planning, but still require additional mechanisms to prune redundant exploration or maintain plan quality at scale. Parallel efforts incorporate generative models to predict unknown regions~\cite{ho2025mapex,wang2025cogniplan} or to guide long-term planning~\cite{cao2025dare}, which alleviates myopic behaviors. Overall, although these methods significantly advance navigation under uncertainty, they operate as heuristic, single-agent policies and do not benefit from team-level cooperation. Moreover, their decision-making is tightly coupled to individual belief updates and is not suitable to handle inter-agent interactions, making them difficult to transfer directly to multi-agent navigation settings.

\subsection{Cooperative Multi-Agent Navigation}
Beyond single-robot navigation, achieving cooperation across a team of robots poses more challenges. Most works in multi-agent navigation focus on guaranteeing collision-free motion at the kinematic level~\cite{jin2021hierarchical,yu2023learning,wang2020mobile}, often in cluttered/crowded environments. These methods primarily treat other agents as local dynamic constraints, and their policies are designed to ensure short-horizon safety and feasibility. While some of them explicitly consider environmental uncertainty~\cite{wagner2017path}, this decision-making is still tightly coupled to individual belief updates, making it difficult to endow robots with long-horizon, team-level reasoning in large, partially known environments. Another line of work studies multi-agent navigation while keeping a free-floating formation of specific shape~\cite{tanner2012multiagent,quan2023robust}. These approaches design behavior-based rules to induce globally consistent motion, but they typically assume known environments, fixed formation objectives, or pre-specified connectivity patterns. Multi-Agent Path Finding (MAPF) offers a more global view of coordination, aiming to compute conflict-free trajectories that minimize makespan. Representative methods include bounded-suboptimal~\cite{li2021eecbs}, prioritized/rule-based search~\cite{li2022mapf}, as well as learning-augmented~\cite{sartoretti2019primal} and lifelong variants~\cite{jiang2025deploying} that scale to thousands of agents. Despite their strong scalability and coordination in discrete, fully known environments, extending MAPF to continuous, partially known environments with online execution remains challenging. 
A few works move closer to cooperative reasoning by studying multi-agent path topologies without explicit communication~\cite{mavrogiannis2019multi}, enabling targeted rendezvous/message passing among agents~\cite{das2019tarmac,tan2024ir,agarwal2025lpac}, or co-optimizing policies with reconfigurable environments~\cite{gao2025co}. Yet, they remain largely heuristics-based, relying on implicit communication that emerges from local observations rather than deliberate intent sharing, and thus lack mechanisms to model temporally extended, altruistic cooperation, such as when to sacrifice individual progress or re-engage to assist teammates.
\begin{figure*}
  \centering
  \includegraphics[width=0.95\textwidth]{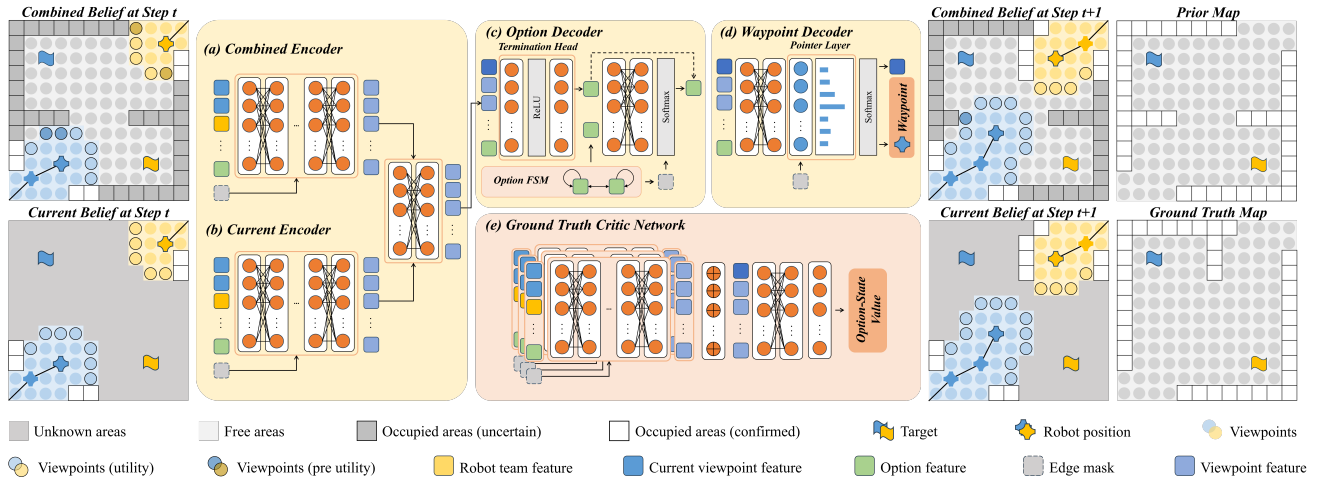}
  \caption{Policy and multi-agent critic networks of ORION. The combined and current encoders fuse prior information with online observations into joint features. A termination head and option decoder then decide whether to maintain the current option or switch to a new valid one, while the waypoint decoder integrates the option feature with the current node feature to select a waypoint from the agent's neighboring nodes. In parallel, the critic network, conditioned on the states, actions, and selected options of all agents, provides centralized value estimates. During training, it leverages the ground-truth map as privileged information to estimate option--state values that capture the long-term team return.}
  \label{fig: orion}
  \vspace{-0.2cm}
\end{figure*}

\section{The ORION Framework}
\subsection{Problem Formulation}

We study multi-agent navigation in partially known environments with online map sharing. A team of $N$ agents $\mathcal{A}=\{1,\dots,N\}$ is initially provided a 2D prior occupancy map $\mathcal{M}^-$ consisting of free regions $\mathcal{M}_f^-$ and occupied regions $\mathcal{M}_o^-$. The true environment is described by a ground truth map $\mathcal{M}_g$, which may partially differ from $\mathcal{M}^-$ due to layout changes, such as shelf rearrangements. During execution, agents share sensor observations through perfect global communication to incrementally build a current map $\mathcal{M}$, which is initially unknown and gradually reveals the true structure of the environment. To reason jointly over prior knowledge and newly revealed areas, agents maintain a combined map $\hat{\mathcal{M}}$, initialized as $\mathcal{M}^-$ and continuously updated using observations accumulated in $\mathcal{M}$. The environment is modeled as a graph $\mathcal{G}=(\mathcal{V},\mathcal{E})$, where nodes $v\in\mathcal{V}$ are uniformly sampled from free space in $\hat{\mathcal{M}}$ (and $\mathcal{M}$) and edges $(v_i, v_j)\in\mathcal{E}$ represent collision-free transitions between neighboring nodes, with each node connected to up to $k$ neighbors. Each agent starts from an initial node $s_i\in\mathcal{V}$ and aims to reach a designated target node $g_i\in\mathcal{V}$. At each decision step, an agent may move to a neighboring node or remain stationary. Agents must avoid \textit{vertex and edge collisions}, i.e., no two agents may occupy the same node at the same time, i.e., $v^i_t \ne v^i_t$ for all $i \neq j$ or traverse the same edge in opposite directions, i.e., $(v^i_t, v^i_{t+1})\ne (v^j_{t+1}, v^j_t)$ for all $i \neq j$. The objective is to generate a set of collision-free trajectories $\Psi=\{\psi_1,\dots,\psi_N\}$ that guide all agents to their targets through sequence of neighboring nodes while minimizing the team makespan, defined as $\min_\Psi \max_{i\in \mathcal{A}}T_i$, where $T_i$ denotes the arrival time (or distance) of agent $i$.

During the task, robots undergo two distinctive stages: During the \emph{pre-arrival} stage, an agent prioritizes navigating toward its target while opportunistically collecting information that may reduce uncertainty for the team. Once it reaches its target, i.e., in the \emph{post-arrival} stage, the agent autonomously decides to either remain at its target until the overall task completes, or temporarily depart to explore nearby uncertain areas and assist teammates still en route, provided it can return in time. 

\subsection{Observation as Informative Graph}
Each node/viewpoint is augmented with semantic attributes beyond 2D coordinates to obtain richer representations: 
(1)~\textbf{Utility $u_j$}: Following~\cite{yamauchi1998frontier,liang2024hdplanner}, frontier measures as boundaries between free space and unknown areas in the current map $\mathcal{M}$, and the utility $u_j$ denotes the number of such frontiers at node $v_j$, which are visible, i.e., along a collision-free line of sight.
(2)~\textbf{Prior utility $u^p_{j}$}: Similar to frontiers, prior frontiers represent the boundary between confirmed free space (validated by the current map) and uncertain obstacles inferred from the prior map. It combines confirmed observations in $\mathcal{M}$ with prior knowledge in $\mathcal{M}^-$. These prior frontiers capture the potential information gain suggested by the prior map, offering a long-range heuristic of current unknown areas. Prior utility $u^p_{j}$ is defined as the number of observable prior frontiers at node $v_j$.
(3)~\textbf{Visited flag $\delta_{j,i}$}: indicates whether $v_j$ has been visited by agent $i$.  
(4)~\textbf{Verified signal $s_j$}: marks whether $v_j$ lies within $\mathcal{M}$.  
(5)~\textbf{Occupancy indicator $p_{j,i}$}: as a ternary variable, encodes whether $v_j$ is currently occupied by agent $i$.
(6)~\textbf{Target binary $t_{j,i}$}: specifies whether $v_j$ coincides with agent $i$'s own target or with another agent's target (if any). We further cluster all viewpoints with non-zero utility within a range $r_b$ into a \emph{beacon set}~\cite{liang2024hdplanner}, representing candidate regions to navigate. We further construct a planning graph $\bar{G}_t=(\bar{V}_t,\bar{E}_t)$ on $\mathcal{M}$, containing samples from both free and unknown cells, where edges connect collision-free pairs of neighbors. Based on $\mathcal{M}$, we define two guideposts that facilitate coordinated navigation:  
(7)~\textbf{Navigation guidepost $g_{i}$}: the Dijkstra trajectory on $\bar{G}_t$ from agent $i$ to the beacon nearest the agent's assigned target~\cite{dijkstra2022note}, which serves as a feasible trajectory cue for self-navigation. 
(8)~\textbf{Cooperation guidepost $g^{c}_{i}$}: if any other target remains unverified, i.e., not in the explored area, we compute the trajectory from agent $i$ to the beacon nearest the closest unverified target; otherwise $g_{i}=g^{c}_{i}$ if all targets are verified. This guidance makes it feasible for agents to assist teammates.

\subsection{Policy and Critic Networks}
\label{subsec:network}
\subsubsection{\textbf{Graph-based Encoders}}

As shown in Figure~\ref{fig: orion}, we design a graph attention-based network that includes an updated and a combined encoder to fuse online, reliable observations from the current map with the rest of the knowledge that may not be up to date from the prior map. The updated encoder processes the current graph $G_t$ into $d$-dimensional feature $\mathbf{s}^n \in \mathbb{R}^{n \times d}$. Specifically, it begins with a feed-forward layer and then applies multiple masked self-attention layers. The mask set $M\in \mathbb{R}^{n \times n}$, obtained from edge connections $E_t$, restricts each node's attention to its neighbors. On the other hand, the combined encoder shares the same architecture as the updated encoder but operates on the combined augmented graph $\hat{G}_t$, producing feature $\mathbf{\hat{s}}^n \in \mathbb{R}^{n \times d}$. Subsequently, we fuse the two feature sets $\mathbf{s}^n$ and $\mathbf{\hat{s}}^n$ via a cross-attention layer into the robot state feature $\mathbf{s} \in \mathbb{R}^d$, enabling the model to selectively integrate prior information by down-weighing inconsistent priors and emphasizing those that align with the updated belief.

\subsubsection{\textbf{Option Decoder} $\pi_{\vartheta}$}

During multi-agent navigation, an agent may either navigate toward its designated target or assist others by collecting more information that benefits the team, each demanding distinct and sometimes conflicting reasoning. In such cases, adaptive switching between self-directed and cooperative behaviors becomes crucial. To balance individual efficiency and collective performance for improved overall task success, we introduce the option-critic structure~\cite{bacon2017option}, where an option denotes a high-level behavioral mode characterized by a termination head function and an intra-option policy, while the critic provides a value-based estimation guiding high-level option/low-level action optimization. Specifically, we augment the policy network with an \emph{option decoder}, which determines whether to continue the current high-level behavior or terminate and switch to another. Given the current state representation feature $\mathbf{s}_t$ and the active option $z_{t-1}$ at the last step, we define a termination head function $\beta_\vartheta(\mathbf{s}_t,z_{t-1})$ that estimates the likelihood of terminating the current option at step $t$. Specifically, we first obtain a joint feature representation by combining the current node feature and the embedding of the previous option $h_t = \mathbf{s}_t+e(z_{t-1}),$ where $e(z_{t-1}) \in \mathbb{R}^d$ denotes the learnable option embedding. 
We parameterize the termination function as $\beta_{\vartheta}(\mathbf{s}_t, z_{t-1}) = \sigma\!\big(\mathrm{MLP}_{\vartheta}(h_t)\big),$ where $\mathrm{MLP}_{\vartheta}(\cdot)$ is a two-layer feed-forward network with ReLU activation, and $\sigma(\cdot)$ denotes the logistic sigmoid.
Finally, a Bernoulli sampling $\mathbf{1}_{\text{term}}/\mathbf{1}_{\text{cont}}\sim\mathrm{Bern}(\beta_\vartheta)$ determines whether the current option should terminate or not.

Although our option framework enables flexible switching between high-level behavioral modes, unconstrained transitions may lead to inconsistent/invalid options. Here we introduce a \emph{finite-state machine} (FSM) that encodes admissible transitions between options at different stages of the task. For a general option set $\mathcal{Z}=\{z_{1},\ldots,z_{K}\}$, we define a binary transition matrix $M(\mathbf{s}_t)\in\{0,1\}^{K\times K},$ where each entry $M_{ij}(\mathbf{s}_t)$ specifies whether transitioning from option $i$ to option $j$ is allowed. This mask captures explicit coordination rules at different stages. The option-selection distribution is therefore computed via a masked softmax:
\begin{equation}
\pi_\vartheta(z_t \mid \mathbf{s}_t, z_{t-1})
=
\frac{
M_{z_{t-1},z_t}(\mathbf{s}_t)\,
\exp\!\big(\beta_{\vartheta}(\mathbf{s}_t,z_t)\big)
}{
\sum_{z'\in\mathcal{Z}}
M_{z_{t-1},z'}(\mathbf{s}_t)\,
\exp\!\big(\beta_{\vartheta}(\mathbf{s}_t,z')\big)
}.
\label{eq:masked_option_policy}
\end{equation}

In our navigation setting, the option space simplifies into two high-level modes, i.e., self-directed navigation and cooperative assistance, yet the FSM remains essential: these two modes play fundamentally different roles before and after the agent reaches its individual target. During the \emph{pre-arrival} phase, both behaviors are permissible, enabling flexible coordination. In contrast, during the \emph{post-arrival} phase, our FSM suppresses self-directed continuation and biases option transitions toward the cooperative mode, reflecting the fact that an agent that has completed its own task should primarily support others. At execution time, the termination decision $\mathbf{1}_{\text{term}}/\mathbf{1}_{\text{cont}}\sim\mathrm{Bern}(\beta_\vartheta)$ determines whether the agent remains committed to the current high-level mode or consults the FSM selector to transition into another valid option:
\[
z_t =
\begin{cases}
z_{t-1}, & \text{if } \mathbf{1}_{\textup{term}}=0,\\[4pt]
z'\sim\pi_\vartheta(\cdot\mid \mathbf{s}_t,z_{t-1}), & \text{if } \mathbf{1}_{\textup{term}}=1.
\end{cases}
\]

Once a high-level option $z_t$ is selected (or continued from the previous step), the intra-option policy is realized through a pointer-network decoder that generates the next waypoint conditioned on both the current node representation and the option embedding. Specifically, we first form the option-conditioned latent feature $\tilde{\mathbf{h}}_t = \mathbf{h}_t + \mathbf{e}(z_t)$, where $\mathbf{h}_t \in \mathbf{s}_t$ denotes the encoded feature of the current node and $\mathbf{e}(z_t)$ is a learned embedding that modulates the decoder according to the current option. This additive conditioning biases the decoder's attention toward option-specific spatial preferences, allowing the model to express distinct behaviors under different options. The decoder then attends over the neighboring node features $H_i = \{\mathbf{h}_j\}_{j \in \mathcal{N}(i)}$, where $\mathcal{N}(i)$ indicates the set of node $v_i$'s neighbors, producing an enhanced node representation $\mathbf{h}_t' = \mathrm{Decoder}(\tilde{\mathbf{h}}_t, H_{i}),$ which integrates both the option-conditioned query and contextual neighbor information via cross-attention.  
Based on this decoded representation, a pointer layer computes attention scores over all candidate waypoints, and finally, the decoder outputs option-specific action logits $\pi_\theta(a_t \mid \mathbf{s}_t, z_t) = \mathrm{softmax}\big(\text{Pointer}(\mathbf{h}_t', H_{i})\big),$ where the selected waypoint $\hat{a}_t = \arg\max_{a_t} \pi_\theta(a_t \mid \mathbf{s}_t, z_t)$ determines the robot's next move.
\subsubsection{\textbf{Multi-Agent Critic Networks}}
For agent $i$, local feature $\mathbf{h}_i$ is extracted from the critic encoder to capture individual observations without inter-agent interaction. Enhanced feature $\tilde{\mathbf{h}}_i$ further aggregates team-wide context via cross-attention layers.
To model coordination, $\tilde{\mathbf{h}}_{i,j}$ denotes the relational embedding between agent $i$ and $j$, capturing their relative states under a shared team belief. 
These features are concatenated to form the local state--action embedding $\mathbf{u}_{i,j} = f_{\mathrm{emb}}([\tilde{\mathbf{h}}_i,\tilde{\mathbf{h}}_{i,j}])$, which jointly integrates the agent's local state, inter-agent relationships, and the global context of the environment. Following MAAC~\cite{iqbal2019actor}, the critic computes a query
$\mathbf{q}_i = W_q\mathbf{h}_i$ and attends to keys and values
$\mathbf{k}_j = W_k\mathbf{u}_{i,j}$, $\mathbf{v}_j = W_v\mathbf{u}_{i,j}$ via
\[
    \alpha_{ij}=\frac{\exp(\mathbf{q}_i^\top\!\mathbf{k}_j)}{\sum_{\ell\neq i}\exp(\mathbf{q}_i^\top\!\mathbf{k}_\ell)},\qquad
    \mathbf{c}_i=\sum_{j\neq i}\alpha_{ij}\mathbf{v}_j.
\]
Finally, the critic forms $Q_{i}=f_Q([\mathbf{u}_{i,j}]_{j\in\mathcal{N}}),$ where $f_Q$ is the embedding layer to output the centralized action-value estimation $Q_i$ for agent $i$ in the robot team $\mathcal{N}$, which integrates local features, multi-agent interactions, and action-specific neighborhood structure. Our critic network adopts the same graph-transformer backbone as the policy network, but differs in its use of a \emph{privileged graph encoder}. During training, this encoder has access to privileged ground-truth information (i.e., the ground-truth map), enabling the critic to provide low-variance estimations on long-term returns to the decentralized policies. Unlike the actor's dual encoders (one for the combined graph and one for the current graph), the critic employs a single unified encoder that captures the complete relational context of all agents and their robot beliefs. Besides the state representation input, the critic also takes the option feature selected at the previous step as input, which benefits both option and waypoint optimization during critic learning~\cite{iqbal2019actor}. The value estimate is explicitly conditioned on the agent's previous option $z_{t-1}$, representing the high-level mode active at the current step. Specifically, the option embedding is added to the current node feature:
\begin{equation}
\mathbf{h}_t^i = \mathrm{Gather}(H_t, \text{index}=i) + e(z_{t-1}^i),
\label{eq:option_embed}
\end{equation}
where $H_t$ is the encoded graph representation and $e(\cdot)$ is an embedding layer. The resulting feature $\mathbf{h}_t^i$ is then refined by a transformer decoder queried with other agents' features, forming an option-aware latent representation. Finally, instead of outputting a waypoint distribution, the critic decoder produces the \emph{option-state values} as the long-term estimation.

Our critic network enables termination learning by contrasting/advantaging the expected return of continuing the current option with that of terminating it. For agent $i$, given the per-agent Q-values $Q_\vartheta^i(\mathbf{s}_t, a_t, z_t)$ and the counterfactual value if the option terminates
$Q_\vartheta^i(\mathbf{s}_t, a_t, \bar{z_t})$, we compute the advantage of termination as $A_{\text{term}}^i = Q_\vartheta^i(\mathbf{s}_t, a_t, \bar{z_t}) - Q_\vartheta^i(\mathbf{s}_t, a_t, z_t),$ which quantifies whether switching options yields a higher expected value than continuing. In the policy network's option decoder, the termination head outputs $\mathbf{1}_{\text{term}}/\mathbf{1}_{\text{cont}}\sim\mathrm{Bern}(\beta_\vartheta)$ through a scalar logit,
from which we define the Bernoulli probabilities of termination and continuation: $p_{\text{term}}^i\sim\log \beta_\vartheta(\cdot\mid \mathbf{s}_t,z_{t-1})$ and $p_{\text{cont}}^i\sim\log \beta_\vartheta(\cdot\mid \mathbf{s}_t,\bar{z}_{t-1})$.
Given the binary termination signal $\delta_t^i \in \{0,1\}$ from the termination head, the likelihood of the observed decision is $p_{\vartheta}^i =
\mathbf{1}_{\text{term}} p_{\text{term}}^i +
\mathbf{1}_{\text{cont}}p_{\text{cont}}^i$.

Overall, we optimize the termination head by maximizing the expected return differential,
i.e., agents are encouraged to terminate when doing so leads to higher option-state values.
The termination head loss is thus defined as $ \mathcal{L}_{\vartheta} = -\,\mathbb{E} \!\left[\log p_{\vartheta}^i \, A_{\text{term}}^i\right]$. On the other hand, following~\cite{haarnoja2018soft}, the policy loss is further improved by our option-regularized policy network:
\begin{align}
\mathcal{L}_{\pi}
&=
\mathbb{E}_{(\mathbf{s}_t,z_t)}
\Bigg[
\sum_{a_t}
\pi_\theta(a_t \mid \mathbf{s}_t,z_t)\,
\Big(
\alpha\,\log \pi_\theta(a_t \mid \mathbf{s}_t,z_t)
\notag\\[-1mm]
&\qquad\qquad\qquad\qquad
-
Q_{\vartheta}(\mathbf{s}_t,a_t,z_t)
\Big)
\Bigg].
\label{eq:oc_policy_loss}
\end{align}

Finally, the overall loss is $\mathcal{L}_{\text{policy}} = \mathcal{L}_{\pi} + \lambda_{\vartheta}\,\mathcal{L}_{\vartheta},$ where $\lambda_{\vartheta}$ balances termination learning. Following~\cite{haarnoja2018soft}, the critic is optimized via a temporal-difference regression objective.

\begin{table*}[ht]
  \caption{Comparisons on \SI{160}{m}$\times$\SI{150}{m} simulated maps. Each entry reports the average value with its variance in parentheses.}
  \label{tab:python}
  \centering
  \resizebox{\textwidth}{!}{%
  \begin{tabular}{lcccccccc}
    \toprule
    \multirow{2}{*}{\textbf{Planner}} &
    \multicolumn{2}{c}{\textbf{3 Agents}} &
    \multicolumn{2}{c}{\textbf{4 Agents}} &
    \multicolumn{2}{c}{\textbf{5 Agents}} &
    \multicolumn{2}{c}{\textbf{10 Agents}} \\
    \cmidrule(lr){2-3}\cmidrule(lr){4-5}\cmidrule(lr){6-7}\cmidrule(lr){8-9}
    & \textbf{Max Dist} (\si{m})$\downarrow$ & \textbf{Steps} $\downarrow$
    & \textbf{Max Dist} (\si{m})$\downarrow$ & \textbf{Steps} $\downarrow$
    & \textbf{Max Dist} (\si{m})$\downarrow$ & \textbf{Steps} $\downarrow$
    & \textbf{Max Dist} (\si{m})$\downarrow$ & \textbf{Steps} $\downarrow$ \\
    \midrule
    MAContext w/o dual-stage~\cite{liang2023context} & 522.07 ($\pm$162.09) & 64.59 & 510.11 ($\pm$170.3) & 62.64 & 495.39 ($\pm$151.08) & 61.03 & 583.66 ($\pm$177.65) & 71.52 \\
    MAContext~\cite{liang2023context}                & 503.3  ($\pm$152.98) & 61.63 & 495.45 ($\pm$129.99) & 61    & 478.49 ($\pm$110.82) & 59.16 & 500.77 ($\pm$168.78) & 60.53 \\
    EECBS~\cite{li2021eecbs}                         & 526.12 ($\pm$151.22) & 66.08 & 501.78 ($\pm$127.77) & 62.64 & 477.12 ($\pm$129.86) & 59.49 & 468.3  ($\pm$136.55) & 58.6 \\
    LNS2~\cite{li2022mapf}                           & 520.99 ($\pm$141.15) & 65.09 & 499.24 ($\pm$119.46) & 62.52 & 474.32 ($\pm$127.82) & 59.49 & 466.24 ($\pm$141.68) & 58.44 \\
    \midrule
    \textbf{ORION} w/o option                                 & 492.83 ($\pm$139.02) & 59.98 & 484.73 ($\pm$120.26) & 58.98 & 467    ($\pm$119.51) & 56.92 & 457.29 ($\pm$129.14) & 55.56 \\
    \textbf{ORION} w/o current                                  & 490.13 ($\pm$135.07) & 60.41 & 470.53 ($\pm$93.45) & 58.42 & 467.88    ($\pm$98.65) & 58.54 & 445.71 ($\pm$107.15) & 55.32 \\
    \textbf{ORION} w/o combined                                  & 500.11 ($\pm$137.87) & 61.45 & 480.81 ($\pm$138.03) & 59.98 & 472.29    ($\pm$133.14) & 58.76 & 468.28 ($\pm$112.92) & 57.6 \\
    \textbf{ORION} w/o dual-stage                             & 462.91 ($\pm$118.24) & 56.08 & 453.85 ($\pm$131.57) & 55.2  & 447.7  ($\pm$125.5)  & 54.54 & 438.31 ($\pm$141.91) & 54.69 \\
    \textbf{ORION}                                            & \textbf{455.83} ($\pm$108.51) & \textbf{55.36} & \textbf{441.99} ($\pm$107.2) & \textbf{53.86} & \textbf{439.24} ($\pm$111.08) & \textbf{53.24} & \textbf{435.76} ($\pm$99.28) & \textbf{53.26} \\
    \bottomrule
  \end{tabular}}
  \vspace{-6pt}
\end{table*}

\begin{table*}[ht]
    \caption{Comparisons on Gazebo simulations. Each entry reports the maximum, average, and minimum travel distance.}
    \label{tab:ros}
    \centering
    \resizebox{\textwidth}{!}{%
    \begin{tabular}{lccccccccc}
        \toprule
        \multirow{2}{*}{\textbf{Planner}} & \multicolumn{3}{c}{\textbf{3 Agents}} & \multicolumn{3}{c}{\textbf{4 Agents}} & \multicolumn{3}{c}{\textbf{5 Agents}} \\
        \cmidrule(lr){2-4}\cmidrule(lr){5-7}\cmidrule(lr){8-10}
        & \textbf{Max Dist} $\downarrow$ & \textbf{Avg Dist} $\downarrow$ & \textbf{Min Dist} $\downarrow$
        & \textbf{Max Dist} $\downarrow$ & \textbf{Avg Dist} $\downarrow$ & \textbf{Min Dist} $\downarrow$
        & \textbf{Max Dist} $\downarrow$ & \textbf{Avg Dist} $\downarrow$ & \textbf{Min Dist} $\downarrow$\\
        \midrule
        MAContext~\cite{liang2023context}    & 178.83 & 110.95 & \textbf{54.55} & 228.13 & 120.53 & 57.78  & 141.98 & 91.5& \textbf{60.75}\\
        \textbf{ORION} w/o option    & 111.74 & 96.6 & 71.45 & 102.46 & 81.86 & 57.02  & 108.7 & 92.26&64.49\\
        \textbf{ORION}               & \textbf{75.99} & \textbf{64.2} & 55.34 & \textbf{72.68} & \textbf{63.5} & \textbf{54.14} & \textbf{95.64}& \textbf{75.06}& 64.58\\
        \bottomrule
    \end{tabular}}
\end{table*}
\section{Experiments}

\subsection{Comparisons in Simulated Maps}
We expand the navigation dataset released in~\cite{liang2023context} to train ORION with a sensor range of \SI{20}{m}. For graph construction, we uniformly sample \SI{600}{} points across the environment and treat points in known free space as candidate viewpoints. Each viewpoint is connected to its \SI{20}{} nearest neighbors while retaining only collision-free edges. ORION is trained on a workstation with one i9-10980XE CPU and two NVIDIA GeForce RTX 3090 GPUs. Training typically converges after \SI{60000}{} episodes and takes around one day. Following recent navigation and exploration benchmarks~\cite{liang2023context,tan2024ir}, we further introduce a benchmark of \SI{500}{} large and challenging maze-like simulated maps to evaluate ORION against representative baselines. For the learning-based baseline, we consider \textbf{MAContext}~\cite{liang2023context}, a multi-agent extension of a DRL navigation planner designed for partially known environments. For classical MAPF solvers, we include 1) \textbf{LNS2}~\cite{li2022mapf}, a highly efficient and scalable planner that iteratively repairs collision-containing path sets by replanning for a subset of colliding agents, and 2) \textbf{EECBS}~\cite{li2021eecbs}, a bounded-suboptimal MAPF planner that combines high-level Explicit Estimation Search with low-level focal search. To deploy LNS2 and EECBS in our partially known setting, we align their execution with ORION by using the same graph-construction pipeline: At each decision timestep, they replan on the constructed graph and select the first waypoint of the planned path as the next waypoint, with the same replanning frequency and collision definitions as ORION. To further analyze our model, we additionally evaluate four ablation variants: 1) \textbf{ORION w/o option}, removing our option-critic structure, 2) \textbf{ORION w/o current}, current map being unavailable, 3) \textbf{ORION w/o combined}, excluding the combined map, and 4) \textbf{ORION w/o dual-stage}, disabling our pre-/post-arrival stages. These variants help isolate the role of our proposed option-critic's hierarchical decision-making, dual map representations, and dual-stage cooperation strategy.

\begin{figure*}[t]
  \centering
  \includegraphics[width=0.88\textwidth]{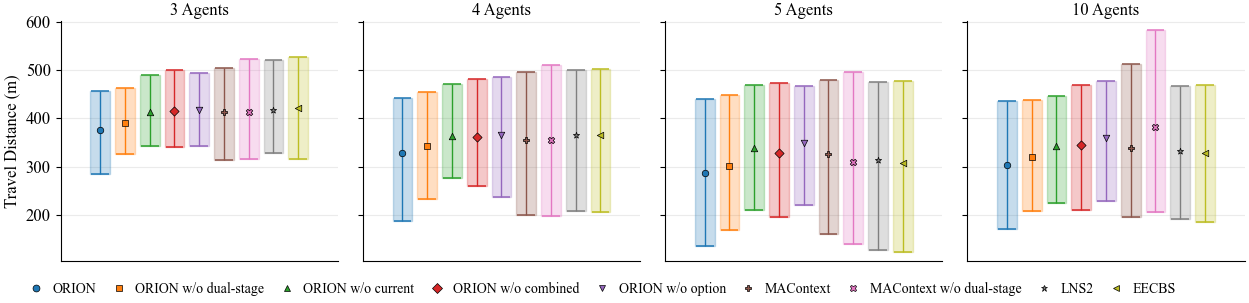}
  \caption{Comparison of travel distances on simulated maps. For each planner, each bar shows the makespan, average travel distance, and minimum team distance as the top, middle, and bottom markers, respectively. Results are reported for teams of $3$, $4$, $5$, and $10$ agents.}
  \label{fig: py}
  \vspace{-0.2cm}
\end{figure*}

As shown in Table~\ref{tab:python}, ORION achieves the best overall performance across all team sizes, reducing makespan by up to \SI{20}{\%} over state-of-the-art planners. Traditional baselines (LNS2 and EECBS) maintain competitive as the team grows; however, ORION consistently outperforms them, exhibiting a \SI{12.5}{}–\SI{13.4}{\%} advantage at 3 agents and still retaining \SI{6.5}{}–\SI{6.9}{\%} improvement at 10 agents. In contrast, the learning-based baseline MAContext degrades with scale, with makespan increasing from \SI{478.5}{m} to \SI{500.8}{m} as the number of agents increases from 5 to 10. ORION, however, maintains stable makespan and decision steps, highlighting stronger scalability and generalization. Although our task objective focuses on minimizing the team's makespan, ORION also improves the average and minimum travel distance (Fig.~\ref{fig: py}), further indicating better overall planning efficiency. Qualitatively, ORION exhibits targeted, timing-aware cooperation: agents with low-cost detours or completed goals often help resolve key uncertainties for bottleneck teammates rather than assisting uniformly across the team. Two ablations further indicate that both maps are necessary. ORION w/o combined, i.e., without prior information, causes a clear makespan drop (\SI{7.5}{}--\SI{9.7}{\%} at 3--5 agents), since agents spend more effort exploring before exploiting prior observations. Removing the current map (ORION w/o current) also degrades performance across all team sizes, showing that up-to-date observations are equally important for correcting prior-map errors. Together, these results confirm that the combined map provides useful environmental priors, while the current map supplies accurate online updates.

We further apply our dual-stage cooperation strategy to both ORION and MAContext by allowing agents that have already reached their targets to continue exploring nearby uncertain areas. This post-arrival assistance consistently improves team performance, with benefits becoming more pronounced as team size increases. For instance, MAContext improves makespan from \SI{583.7}{m} to \SI{500.8}{m} when equipped with the dual-stage strategy, highlighting the importance of leveraging idle agents to resolve remaining team-level uncertainties.
\begin{figure}[t]
  \centering
  \includegraphics[width=0.92\linewidth]{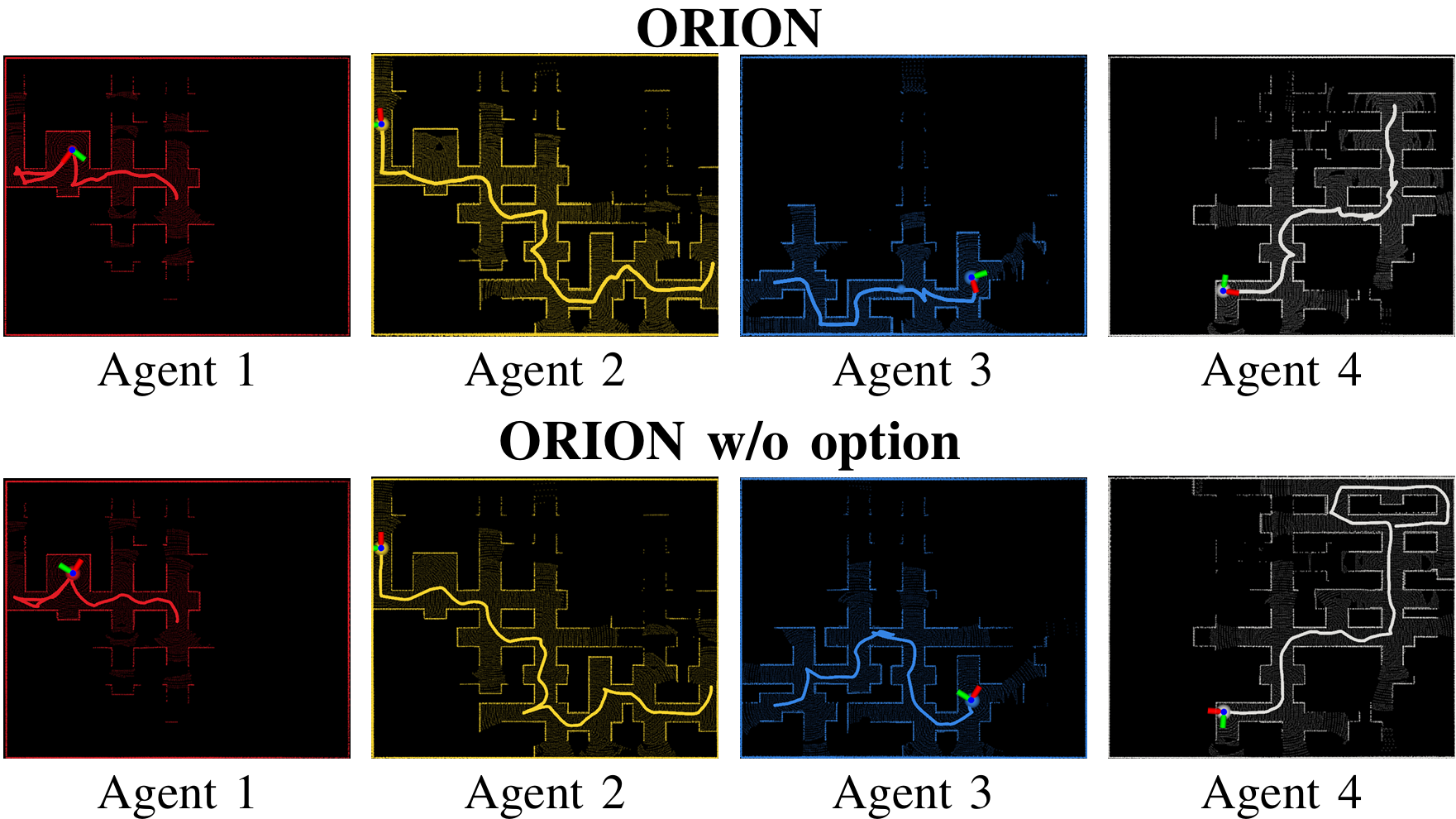}
  \caption{Gazebo Experiments. ORION yields more efficient decentralized coordination than ORION w/o option, as shown by the local maps of the four agents.}
  \label{fig:ros_experiments}
  \vspace{-0.2cm}
\end{figure}
\begin{figure}[t]
  \centering
  \includegraphics[width=0.92\linewidth]{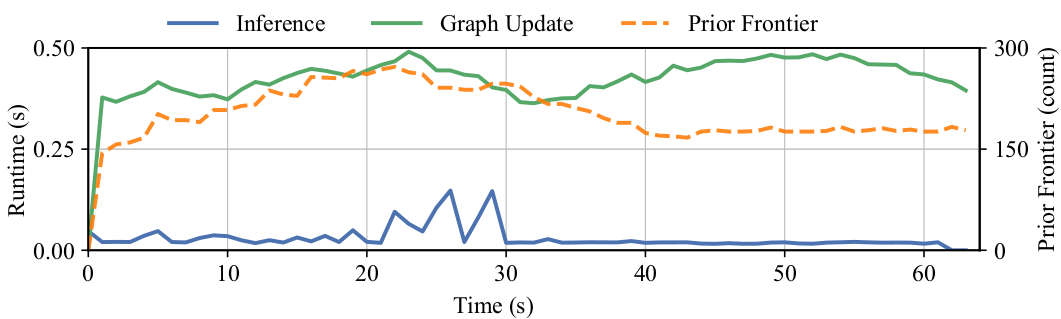}
  \caption{Runtime performance in Gazebo simulation. ORION maintains real-time performance on graph updates and network inference throughout execution, while the prior frontier curve reflects how ORION incrementally verifies and corrects uncertain regions in the prior map during online navigation.}
  \label{fig: ros_curve}
  \vspace{-0.2cm}
\end{figure}
\begin{figure*}[t]
  \centering
  \includegraphics[width=0.92\linewidth]{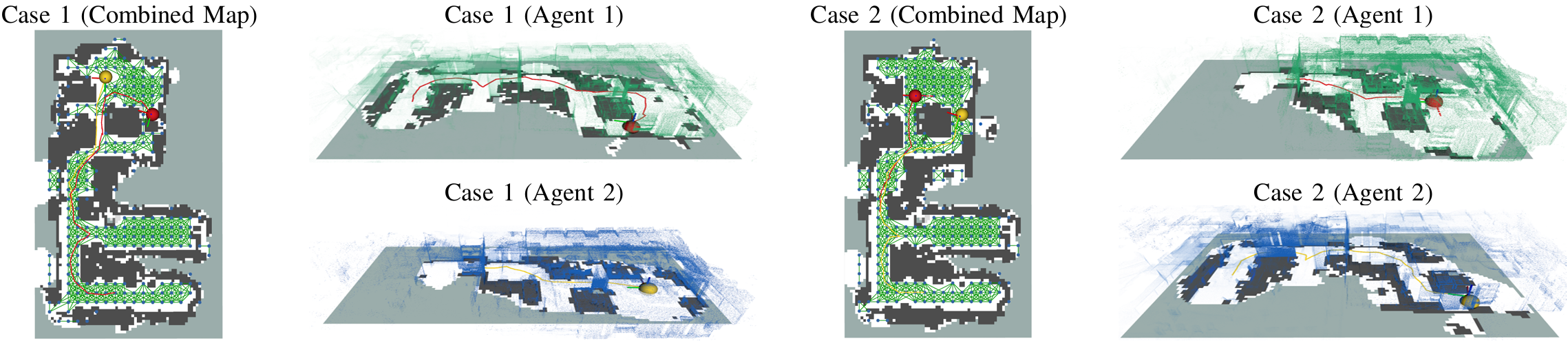}
  \caption{Real-world experiments under two start--target settings. The left column shows the combined map and trajectories, while the right column shows each agent's current belief and executed path.}
  \label{fig: real robot}
  \vspace{-0.2cm}
\end{figure*}
\subsection{Validation in Large-Scale Gazebo Simulations}

We further evaluate our approach in a \SI{70}{m}$\times$\SI{60}{m} Gazebo warehouse environment, where shelf rearrangements cause partial discrepancies between the prior map and the ground truth, thereby simulating realistic online navigation under map changes. ORION and MAContext are deployed on ground vehicle teams (maximum speed \SI{2}{m/s}) equipped with Velodyne VLP16 LiDARs. As shown in Table~\ref{tab:ros}, we compare ORION, ORION w/o option-critic, and MAContext with teams of 3, 4, and 5 robots. ORION consistently achieves the best performance in terms of maximum and average distance. We also observe that, with the option module enabled, ORION brings a performance improvement of \SI{12}{}--\SI{32}{\%} in maximum distance and \SI{19}{}--\SI{34}{\%} in average distance, which further highlights the importance of the option module. While ORION exhibits a slightly larger minimum distance for the largest team, this reflects deliberate post-arrival exploration. Agents that reach their targets early, or are less affected by map changes, take short detours to reduce uncertainty for slower teammates. This trade-off slightly increases individual distance but provides larger team-level gains. It emerges naturally from the option-critic formulation and team-level objective rather than hard-coded heuristics.
 
Fig.~\ref{fig:ros_experiments} illustrates a representative cooperative behavior enabled by ORION: agent~1 reaches its target first (red trajectory) and then moves toward the vicinity of target~2 to share informative observations with agent~2. Upon completing this assistance, agent~1 returns to its own target instead of performing unnecessary exploration, highlighting adaptive cooperation enabled by the option-critic framework and the proposed dual-stage strategy. As shown in Fig.~\ref{fig: ros_curve}, ORION maintains real-time performance throughout the experiments: graph update remains below \SI{0.5}{s}, and network inference requires less than \SI{0.2}{s} (typically under \SI{0.1}{s}), despite the entire system being implemented in Python. We also report the evolution of prior utility, defined as the number of sampled nodes along the boundary between confirmed free space and uncertain obstacles. The prior utility stabilizes after approximately \SI{40}{s}, indicating that the team has obtained a sufficiently confident map and no longer needs extra exploration.

\subsection{Real-World Experiments}

Finally, we deploy ORION on two ground vehicles (maximum speed \SI{0.5}{m/s}) equipped with Livox Mid-360 3D LiDARs for odometry and mapping, operating in a \SI{30}{m}$\times$\SI{10}{m} office environment. We use OctoMap with an occupancy resolution of \SI{0.4}{m} and a waypoint resolution of \SI{0.5}{m} to construct the online map. During online execution, all robots share their locally built current maps, allowing the system to periodically merge them into a global current map. The unknown regions in this global current map are filled with prior-map information, producing the combined map used by all robots. As shown in Fig.~\ref{fig: real robot}, we evaluate ORION under two different start–target configurations that showcase distinct cooperative behaviors enabled by our framework. In both settings, the team travels approximately \SI{20}{m} within \SI{90}{s}. In case~1, agent~1 (red) benefits from areas previously explored by agent~2 (yellow), demonstrating \emph{passive map sharing}, where information gathered by one agent directly assists another without explicit coordination. In case~2, agent~1 exhibits \emph{post-arrival cooperation}: after reaching its target, it moves toward the vicinity of target~2 to provide updated observations. Once target~2 enters the shared belief map, agent~2 immediately returns to its own target without redundant exploration, maintaining a compact travel distance while acquiring critical, up-to-date environmental information. Together, these real-world results highlight ORION's ability to coordinate through both passive map sharing and active post-arrival assistance, underscoring its practicality for multi-robot deployment.

\section{Conclusion}

In this paper, we presented ORION, a deep reinforcement learning framework for cooperative multi-agent online navigation in partially changed environments. By fusing prior maps with online perception through a shared graph encoder and leveraging a novel option-critic structure, ORION enables agents to switch adaptively between target-directed navigation and cooperative assistance in a decentralized way. Our dual-stage cooperation strategy further enhances team-level performance by allowing arrived agents to contribute additional information that helps teammates reach their targets more efficiently. Extensive experiments on large-scale maze-like maps, high-fidelity Gazebo simulations, and physical robot deployments demonstrate that ORION achieves robust, real-time cooperation and consistently outperforms state-of-the-art baselines across varying team sizes. A current limitation lies in its assumption of reliable map sharing among agents. In future work, we plan to explicitly investigate more realistic settings with noisy communication/imperfect localization, where agents must reason about uncertain teammates' beliefs and establish adaptive map-sharing and coordination strategies.

\bibliographystyle{IEEEtran} 
\bibliography{ref}

@inproceedings{yamauchi1998frontier,
  title={Frontier-based exploration using multiple robots},
  author={Yamauchi, Brian},
  booktitle={Proceedings of the second international conference on Autonomous agents},
  pages={47--53},
  year={1998}
}

@inproceedings{iqbal2019actor,
  title={Actor-attention-critic for multi-agent reinforcement learning},
  author={Iqbal, Shariq and Sha, Fei},
  booktitle={International conference on machine learning},
  pages={2961--2970},
  year={2019},
  organization={PMLR}
}

@inproceedings{haarnoja2018soft,
  title={Soft actor-critic: Off-policy maximum entropy deep reinforcement learning with a stochastic actor},
  author={Haarnoja, Tuomas and Zhou, Aurick and Abbeel, Pieter and Levine, Sergey},
  booktitle={International conference on machine learning},
  pages={1861--1870},
  year={2018},
  organization={Pmlr}
}

@inproceedings{das2019tarmac,
  title={Tarmac: Targeted multi-agent communication},
  author={Das, Abhishek and Gervet, Th{\'e}ophile and Romoff, Joshua and Batra, Dhruv and Parikh, Devi and Rabbat, Mike and Pineau, Joelle},
  booktitle={International Conference on machine learning},
  pages={1538--1546},
  year={2019},
  organization={PMLR}
}

@article{mavrogiannis2019multi,
  title={Multi-agent path topology in support of socially competent navigation planning},
  author={Mavrogiannis, Christoforos I and Knepper, Ross A},
  journal={The International Journal of Robotics Research},
  volume={38},
  number={2-3},
  pages={338--356},
  year={2019},
  publisher={SAGE Publications Sage UK: London, England}
}

@article{gao2025co,
  title={Co-Optimizing Reconfigurable Environments and Policies for Decentralized Multi-Agent Navigation},
  author={Gao, Zhan and Yang, Guang and Prorok, Amanda},
  journal={IEEE Transactions on Robotics},
  year={2025},
  publisher={IEEE}
}

@article{quan2023robust,
  title={Robust and efficient trajectory planning for formation flight in dense environments},
  author={Quan {et al.}, Lun},
  journal={IEEE Transactions on Robotics},
  volume={39},
  number={6},
  pages={4785--4804},
  year={2023},
  publisher={IEEE}
}

@article{jin2021hierarchical,
  title={Hierarchical and stable multiagent reinforcement learning for cooperative navigation control},
  author={Jin, Yue and Wei, Shuangqing and Yuan, Jian and Zhang, Xudong},
  journal={IEEE Transactions on Neural Networks and Learning Systems},
  volume={34},
  number={1},
  pages={90--103},
  year={2021},
  publisher={IEEE}
}

@inproceedings{yu2023learning,
  title={Learning control admissibility models with graph neural networks for multi-agent navigation},
  author={Yu, Chenning and Yu, Hongzhan and Gao, Sicun},
  booktitle={Conference on robot learning},
  pages={934--945},
  year={2023},
  organization={PMLR}
}

@article{tanner2012multiagent,
  title={Multiagent navigation functions revisited},
  author={Tanner, Herbert G and Boddu, Adithya},
  journal={IEEE Transactions on Robotics},
  volume={28},
  number={6},
  pages={1346--1359},
  year={2012},
  publisher={IEEE}
}

@inproceedings{jiang2025deploying,
  title={Deploying ten thousand robots: Scalable imitation learning for lifelong multi-agent path finding},
  author={Jiang, He and Wang, Yutong and Veerapaneni, Rishi and Duhan, Tanishq and Sartoretti, Guillaume and Li, Jiaoyang},
  booktitle={2025 IEEE International Conference on Robotics and Automation (ICRA)},
  pages={1--7},
  year={2025},
  organization={IEEE}
}

@article{sartoretti2019primal,
  title={Primal: Pathfinding via reinforcement and imitation multi-agent learning},
  author={Sartoretti {et al.}, Guillaume},
  journal={IEEE Robotics and Automation Letters},
  volume={4},
  number={3},
  pages={2378--2385},
  year={2019},
  publisher={IEEE}
}

@inproceedings{cao2025dare,
  title={Dare: Diffusion policy for autonomous robot exploration},
  author={Cao, Yuhong and Lew, Jeric and Liang, Jingsong and Cheng, Jin and Sartoretti, Guillaume},
  booktitle={2025 IEEE International Conference on Robotics and Automation (ICRA)},
  pages={11987--11993},
  year={2025},
  organization={IEEE}
}

@inproceedings{wang2025cogniplan,
  title={CogniPlan: Uncertainty-Guided Path Planning with Conditional Generative Layout Prediction},
  author={Wang, Yizhuo and He, Haodong and Liang, Jingsong and Cao, Yuhong and Chakraborty, Ritabrata and Sartoretti, Guillaume Adrien},
  booktitle={Conference on Robot Learning},
  pages={1382--1396},
  year={2025},
  organization={PMLR}
}

@inproceedings{ho2025mapex,
  title={Mapex: Indoor structure exploration with probabilistic information gain from global map predictions},
  author={Ho {et al.}, Cherie},
  booktitle={2025 IEEE International Conference on Robotics and Automation (ICRA)},
  pages={13074--13080},
  year={2025},
  organization={IEEE}
}

@inproceedings{wagner2017path,
  title={Path planning for multiple agents under uncertainty},
  author={Wagner, Glenn and Choset, Howie},
  booktitle={Proceedings of the International Conference on Automated Planning and Scheduling},
  volume={27},
  pages={577--585},
  year={2017}
}

@inproceedings{li2022mapf,
  title={MAPF-LNS2: Fast repairing for multi-agent path finding via large neighborhood search},
  author={Li, Jiaoyang and Chen, Zhe and Harabor, Daniel and Stuckey, Peter J and Koenig, Sven},
  booktitle={Proceedings of the AAAI Conference on Artificial Intelligence},
  volume={36},
  number={9},
  pages={10256--10265},
  year={2022}
}

@inproceedings{tan2024ir,
  title={Ir 2: Implicit rendezvous for robotic exploration teams under sparse intermittent connectivity},
  author={Tan, Derek Ming Siang and Ma, Yixiao and Liang, Jingsong and Chng, Yi Cheng and Cao, Yuhong and Sartoretti, Guillaume},
  booktitle={2024 IEEE/RSJ International Conference on Intelligent Robots and Systems (IROS)},
  pages={13245--13252},
  year={2024},
  organization={IEEE}
}

@inproceedings{bacon2017option,
  title={The option-critic architecture},
  author={Bacon, Pierre-Luc and Harb, Jean and Precup, Doina},
  booktitle={Proceedings of the AAAI conference on artificial intelligence},
  volume={31},
  number={1},
  year={2017}
}

@inproceedings{li2021eecbs,
  title={Eecbs: A bounded-suboptimal search for multi-agent path finding},
  author={Li, Jiaoyang and Ruml, Wheeler and Koenig, Sven},
  booktitle={Proceedings of the AAAI conference on artificial intelligence},
  volume={35},
  number={14},
  pages={12353--12362},
  year={2021}
}

@inproceedings{liang2023context,
  title={Context-aware deep reinforcement learning for autonomous robotic navigation in unknown area},
  author={Liang, Jingsong and Wang, Zhichen and Cao, Yuhong and Chiun, Jimmy and Zhang, Mengqi and Sartoretti, Guillaume Adrien},
  booktitle={Conference on Robot Learning},
  pages={1425--1436},
  year={2023},
  organization={PMLR}
}

@article{koenig2005fast,
  title={Fast replanning for navigation in unknown terrain},
  author={Koenig, Sven and Likhachev, Maxim},
  journal={IEEE transactions on robotics},
  volume={21},
  number={3},
  pages={354--363},
  year={2005},
  publisher={IEEE}
}

@article{karaman2011sampling,
  title={Sampling-based algorithms for optimal motion planning},
  author={Karaman, Sertac and Frazzoli, Emilio},
  journal={The international journal of robotics research},
  volume={30},
  number={7},
  pages={846--894},
  year={2011},
  publisher={Sage Publications Sage UK: London, England}
}

@inproceedings{likhachev2005anytime,
  title={Anytime dynamic A*: An anytime, replanning algorithm.},
  author={Likhachev, Maxim and Ferguson, David I and Gordon, Geoffrey J and Stentz, Anthony and Thrun, Sebastian},
  booktitle={ICAPS},
  volume={5},
  pages={262--271},
  year={2005}
}

@inproceedings{gammell2015batch,
  title={Batch informed trees (BIT*): Sampling-based optimal planning via the heuristically guided search of implicit random geometric graphs},
  author={Gammell, Jonathan D and Srinivasa, Siddhartha S and Barfoot, Timothy D},
  booktitle={2015 IEEE international conference on robotics and automation (ICRA)},
  pages={3067--3074},
  year={2015},
  organization={IEEE}
}

@inproceedings{pfeiffer2017perception,
  title={From perception to decision: A data-driven approach to end-to-end motion planning for autonomous ground robots},
  author={Pfeiffer, Mark and Schaeuble, Michael and Nieto, Juan and Siegwart, Roland and Cadena, Cesar},
  booktitle={2017 ieee international conference on robotics and automation (icra)},
  pages={1527--1533},
  year={2017},
  organization={IEEE}
}

@inproceedings{tai2017virtual,
  title={Virtual-to-real deep reinforcement learning: Continuous control of mobile robots for mapless navigation},
  author={Tai, Lei and Paolo, Giuseppe and Liu, Ming},
  booktitle={2017 IEEE/RSJ international conference on intelligent robots and systems (IROS)},
  pages={31--36},
  year={2017},
  organization={IEEE}
}

@inproceedings{jin2020mapless,
  title={Mapless navigation among dynamics with social-safety-awareness: a reinforcement learning approach from 2d laser scans},
  author={Jin, Jun and Nguyen, Nhat M and Sakib, Nazmus and Graves, Daniel and Yao, Hengshuai and Jagersand, Martin},
  booktitle={2020 IEEE international conference on robotics and automation (ICRA)},
  pages={6979--6985},
  year={2020},
  organization={IEEE}
}

@article{liang2024hdplanner,
  title={Hdplanner: Advancing autonomous deployments in unknown environments through hierarchical decision networks},
  author={Liang, Jingsong and Cao, Yuhong and Ma, Yixiao and Zhao, Hanqi and Sartoretti, Guillaume},
  journal={IEEE Robotics and Automation Letters},
  volume={10},
  number={1},
  pages={256--263},
  year={2024},
  publisher={IEEE}
}

@incollection{dijkstra2022note,
  title={A note on two problems in connexion with graphs},
  author={Dijkstra, Edsger W},
  booktitle={Edsger Wybe Dijkstra: his life, work, and legacy},
  pages={287--290},
  year={2022}
}

@article{wang2020mobile,
  title={Mobile robot path planning in dynamic environments through globally guided reinforcement learning},
  author={Wang, Binyu and Liu, Zhe and Li, Qingbiao and Prorok, Amanda},
  journal={IEEE Robotics and Automation Letters},
  volume={5},
  number={4},
  pages={6932--6939},
  year={2020},
  publisher={IEEE}
}

@article{agarwal2025lpac,
  title={Lpac: Learnable perception-action-communication loops with applications to coverage control},
  author={Agarwal, Saurav and Muthukrishnan, Ramya and Gosrich, Walker and Kumar, Vijay and Ribeiro, Alejandro},
  journal={IEEE Transactions on Robotics},
  year={2025},
  publisher={IEEE}
}

\vfill

\end{document}